\documentclass[letterpaper, 10 pt, conference]{ieeeconf}  

\IEEEoverridecommandlockouts                              

\usepackage{cite}
\usepackage{amsmath,amssymb,amsfonts}
\usepackage{algorithmic}
\usepackage{graphicx}
\usepackage{textcomp}
\usepackage{xcolor}

\usepackage{booktabs}
\usepackage{multirow}
\usepackage{graphicx}
\usepackage{amsmath}
\usepackage{url}

\usepackage{wrapfig} 

\def\BibTeX{{\rm B\kern-.05em{\sc i\kern-.025em b}\kern-.08em
    T\kern-.1667em\lower.7ex\hbox{E}\kern-.125emX}}

\title{\LARGE \bf

A Cooperative Contactless Object Transport with Acoustic Robots
}

\author{Narsimlu Kemsaram, Akin Delibasi, James Hardwick, Bonot Gautam, \\Diego Martinez Plasencia, and Sriram Subramanian
\thanks{Narsimlu Kemsaram ({\tt\small n.kemsaram@ucl.ac.uk}), Akin Delibasi ({\tt\small a.delibasi@ucl.ac.uk}), James Hardwick ({\tt\small james.hardwick.19@ucl.ac.uk}), Bonot Gautam ({\tt\small b.gautam@ucl.ac.uk}), Diego Martinez Plasencia ({\tt\small d.plasencia@ucl.ac.uk}), Sriram Subramanian ({\tt\small s.subramanian@ucl.ac.uk}) are with the Department of Computer Science, University College London, United Kingdom.}%
}

\begin{document}

\maketitle
\thispagestyle{empty}
\pagestyle{empty}

\begin{abstract}

Cooperative transport, the simultaneous movement of an object by multiple agents, has been widely observed in biological systems such as ant colonies, which improve efficiency and adaptability in dynamic environments. 
Inspired by these natural phenomena, we present a novel acoustic robotic system for the transport of contactless objects in mid-air. 
Our system leverages phased ultrasonic transducers and a robotic control system onboard to generate localized acoustic pressure fields, enabling precise manipulation of airborne particles and robots. 
We categorize contactless object-transport strategies into independent transport (uncoordinated) and forward-facing cooperative transport (coordinated), drawing parallels with biological systems to optimize efficiency and robustness.
The proposed system is experimentally validated by evaluating levitation stability using a microphone in the measurement lab, transport efficiency through a phase-space motion capture system, and clock synchronization accuracy via an oscilloscope. 
The results demonstrate the feasibility of both independent and cooperative airborne object transport.
This research contributes to the field of acoustophoretic robotics, with potential applications in contactless material handling, micro-assembly, and biomedical applications.

\end{abstract}

\section{Introduction}

Cooperative transport, in which multiple agents collaborate to move an object, is a well-studied phenomenon in biological systems, particularly in ant colonies \cite{wilson1971insect}, \cite{holldobler1990ants}.
In these colonies, individuals take on specialized tasks such as foraging, prey retrieval, and chain formation (see Figure 1A), demonstrating remarkable coordination and adaptability. 
Inspired by this natural behavior, robotic swarms should similarly be able to respond to diverse tasks in different environments by employing appropriate coordination strategies.

\begin{figure}[t!]
    \centering \includegraphics[width=0.95\linewidth]{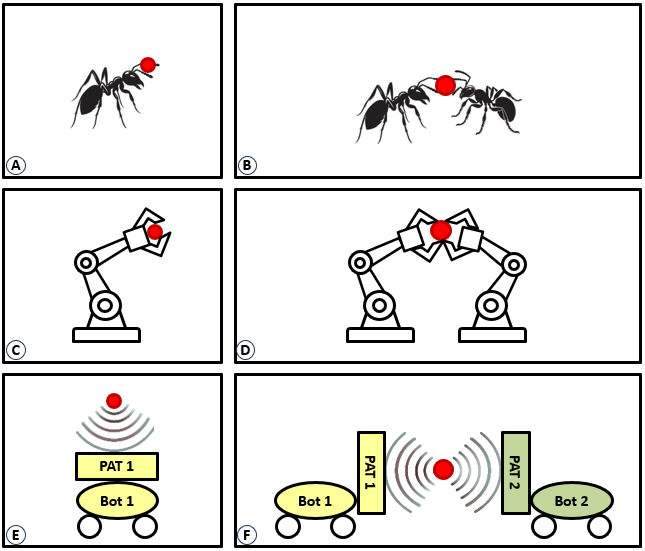}
    \caption{(A)-(B) An independent object transport and a cooperative object transport in ant colonies (contact object transport) \cite{czaczkes2013cooperative}, (C)-(D) An independent object transport and a cooperative object transport in swarm robotics  (contact object transport) \cite{martinoli2005collective}, (E)-(F) Proposed independent and cooperative contactless object transport in the air with acoustic robots (contactless object transport).}
    \label{BlockDiagram}
\end{figure}

In coordinate transport, a group of robots need to cooperate in order to transport an object that is heavy to move by a single robot (see Figure \ref{BlockDiagram}B). 
This process demands synchronized movements and, when necessary, collective decision making, such as resolving stagnation scenarios. 
Drawing inspiration from ants, which efficiently and successfully transport objects through collaboration, robot swarms can develop effective cooperative transport mechanisms, enhancing their ability to navigate and operate in dynamic environments.
Studies \cite{camazine2020self}, \cite{czaczkes2013cooperative} have shown that social insects achieve synchronized operations without relying on a centralized coordination mechanism, but their systems remain robust, scalable, and flexible.
These characteristics are highly desirable for multi-robot systems, making them a key inspiration for swarm robotics.

Biological strategies have inspired significant advancements in swarm robotics, leading to the development of independent transport (see Figure \ref{BlockDiagram}C) and cooperative transport (see Figure \ref{BlockDiagram}D), which have been explored in various robotic systems \cite{martinoli2005collective}, \cite{tuci2018cooperative}.
Early research on swarm-based robotics focused on key aspects such as force distribution, role allocation, and communication strategies for multi-robot object transport \cite{gross2009towards}, \cite{dorigo2021swarm}.
Building on this foundation, later studies demonstrated how control algorithms can enable collective transport without explicit inter-agent communication while ensuring robustness and scalability \cite{rubenstein2012kilobot}, \cite{alkilabi2017cooperative}. 
Although robotic systems have successfully achieved contact-based object transport, achieving the same task in midair or without physical contact remains an open challenge.

Recent advances in acoustic levitation and manipulation have enabled contactless object control using a phased array of transducers \cite{andrade2020acoustic}, \cite{hirayama2022high}. 
This acoustic levitation is based on pressure nodes formed by standing waves, which allow small particles to be suspended and moved within a three-dimensional space.
Previous studies have demonstrated the effectiveness of phased ultrasonic transducers in generating stable acoustic pressure nodes, enabling precise airborne particle manipulation \cite{marzo2015holographic}, \cite{hirayama2019volumetric}. 
However, most existing systems focus on particle control using static fixed units, which lack real-time adaptability and synchronized multi-agent coordination.
This limitation restricts their practical applications, highlighting the need for swarm-based acoustic robots capable of collectively transporting objects in mid-air without physical contact.

In this paper, we present a novel contactless object transport system that utilizes a swarm of acoustic robots to move airborne particles. 
Inspired by biological transport behaviors, we categorize transport strategies into two scenarios: independent transport, where a single robot operates without coordination (see Figure \ref{BlockDiagram}E), and cooperative transport, where multiple robots synchronize their actions to achieve stable transport (see Figure \ref{BlockDiagram}F).
To achieve stable independent and cooperative transport, we develop control strategies that enable a phased array of transducers, dynamically adjusting both the acoustic field and the robot formation control. 
Our system is experimentally validated to assess its robustness and efficiency in real-world scenarios. 

The key contributions of this work are as follows:
i) We introduce a midair object transport system using acoustic robots, enabling both independent transport, where a single robot maintains stable levitation of an object, and cooperative transport, where two robots synchronize their movements to transport an object while walking backward,
ii) We implemented an FPGA- and IR-based clock synchronization mechanism to achieve microsecond-level phase alignment between the leader and follower robots. This precise synchronization is essential to maintain stable acoustic levitation and to ensure accurate contactless object transport, and
iii) We conducted real-world experiments to evaluate the feasibility and effectiveness of our approach, demonstrating the potential of acoustic robots for controlled, contactless transport in midair.

\section{Related Work}

Cooperative transport has been widely explored in both biological systems and robotics, leading to significant advancements in swarm intelligence and multi-agent systems. 
In this section, we review the relevant literature in three key areas: i) biological inspirations for cooperative transport, highlighting natural mechanisms that inform robotic design, ii) swarm-based robotic transport, examining strategies used in multi-robot systems, and iii) acoustic levitation and manipulation, which provide the foundation for contactless object transport.

\subsection{Biological Inspirations for Cooperative Transport}

Cooperative transport is a common phenomenon in social insects \cite{berman2011study}, particularly in ant colonies \cite{berman2011experimental}, where ants employ three primary transport strategies: uncoordinated transport, encircling coordinated transport, and forward-facing coordinated transport \cite{czaczkes2013cooperative}.
In uncoordinated transport, individuals apply force in different directions, resulting in inefficient movement, slow progress, and frequent deadlocks. 
In contrast, encircling coordinated transport - observed in species such as Aphaenogaster ants - involves synchronized movement and balanced load distribution, minimizing conflicts and improving efficiency \cite{mccreery2014cooperative}.
Coordinated forward-facing transport, as seen in army ants, is based on role differentiation, with larger ants leading the transport while smaller ants provide additional support, ensuring a structured and effective transport process \cite{feinerman2018physics}.

These natural transport mechanisms have inspired robotic swarm coordination, as they provide scalable and efficient solutions for object transport. 
By drawing inspiration from biological systems, researchers have developed cooperative object transport strategies that enhance the robustness and adaptability of robotic swarms, enabling them to operate effectively in dynamic environments.

\subsection{Cooperative Transport in Swarm Robotic Systems}

Biological transport strategies have inspired significant advances in swarm robotics, leading to extensive research on cooperative object transport in robotic systems. 
Generally, robotic transport methods can be categorized into three main strategies: pushing, pulling, and caging.
In the push strategy, behavior-based models are used to control robots as they work together to move an object \cite{kube2000cooperative}.
However, this approach presents challenges such as stagnation, motion coordination, and sensitivity to the shape of the transported object, which can hinder efficiency \cite{rubenstein2013collective}.
Pulling transport, on the other hand, involves robots physically attaching themselves to the object using specialized mechanisms.
In ants, pulling an object is the most common way of transport, as ants are equipped with the appropriate physical mechanisms to perform the task \cite{tuci2006cooperation}. 
However, replicating this method in robotics remains challenging due to the complexity of designing suitable mechanical attachments to pull an object \cite{grobeta2008evolution}.
The caging strategy involves multiple robots surrounding an object to enclose it within their formation, effectively trapping it for transport \cite{wang2003control}.
Although this method offers strong control over object movement, it becomes complex when dealing with irregularly shaped objects, as it requires a sufficient number of robots and detailed information about the object's geometry. 

In swarm robotics, cooperative object transport presents a greater challenge due to the need for precise synchronization, stable movement, and the ability to dynamically adapt to external disturbances.
Research into multi-robot transport systems has investigated methods such as tethered payloads, where robots coordinate their trajectory to move objects. 
However, these approaches rely on physical contact with the objects, which limits their scalability, particularly for tasks requiring fine-grained manipulation and transportation.


\subsection{Acoustic Levitation and Manipulation}

Acoustic levitation enables precise contactless manipulation of small particles, and has garnered significant attention for its applications in fields such as material handling, biomedicine, and micro-assembly \cite{stevens1899text}.
Acoustic levitation utilizes standing waves generated by ultrasonic transducers to create pressure nodes that can trap objects in midair, enabling their manipulation without physical contact \cite{foresti2013acoustophoretic}. 
Recent advances have focused on using phased array of transducers, which allow precise control over particles by allowing translation, rotation, and trapping of small objects \cite{melde2016holograms}. 
These developments offer enhanced control and versatility in the manipulation of objects in mid-air, opening new possibilities for contactless applications in science and engineering.
However, cooperative transport using multiple acoustic sources remains an emerging area of research.
Although acoustic tweezers have been successfully developed for precise control at the microscale, large-scale cooperative transport of objects in midair using distributed phased arrays has yet to be fully realized. 

Our work builds on previous research in cooperative transport, swarm robotics, and acoustic manipulation. 
Unlike traditional cooperative transport methods, our proposed system utilizes a swarm of acoustic robots to achieve contactless cooperative object transport. 
By extending biologically inspired transport strategies to an acoustophoretic framework, we introduce a control strategy that enables synchronized transportation using a movable ultrasonic phased array of transducers. 
This approach bridges the gap between swarm robotics and acoustic levitation, offering a novel solution for both independent and cooperative contactless object transport. 
The potential applications of this work are vast, including non-invasive material handling, microfluidics, and biomedical engineering, where precise control and transport are critical.

\section{Concept, Design, Modeling, and Methods} \label{Sec_Design}

\begin{figure}[!htb]
    \centering \includegraphics[width=0.95\linewidth]{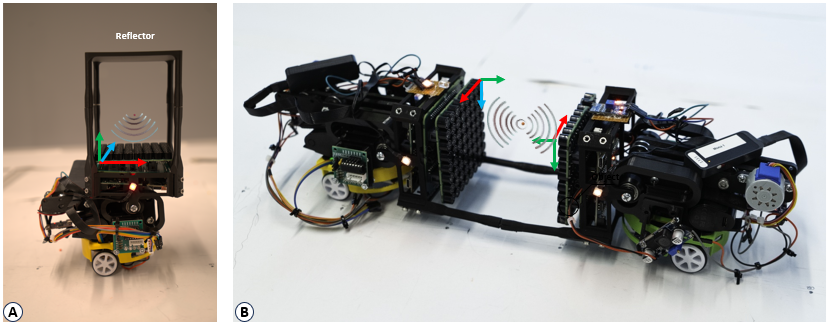}
    \caption{A contactless object transport in mid-air with acoustic robots. (A) An independent contactless transport, where a single acoustic robot moves an object without physical contact, and (B) A cooperative contactless transport, where a swarm of acoustic robots collaboratively moves an object in mid-air.}
    \label{ProposedMethodology}
\end{figure}

Inspired by object transport mechanisms in ants and robotic systems, our approach utilizes acoustic robots equipped with phased arrays to generate dynamic acoustic fields.
These fields enable both independent and cooperativeative contactless transport of airborne particles by a single acoustic robot or a swarm.
Object transport is achieved without physical contact through the generation of acoustic fields and the implementation of robot control algorithms. 
In this work, custom-designed acoustic robots create a levitation zone, allowing objects to stably levitate in midair while the robots move. 
By adjusting the phase difference between the acoustic emitters, these levitation zones can be moved axially, either vertically for independent object transport or horizontally for cooperative transport scenarios (see Figure \ref{ProposedMethodology}).

\subsection{Independent Contactless Object Transport}

In this scenario, an acoustic robot is equipped with a phased array that generates precise single-sided acoustic fields to manipulate and transport an object in midair. 
In addition, a differential drive control system ensures that the robot maintains stable levitation during contactless object transport (see Figure \ref{ProposedMethodology}A).

\subsubsection{Single-sided acoustic field generation}

The generation of single-sided acoustic fields is the most challenging because of the absence of sound waves propagating in the opposite direction.
However,  we can still create a vortex trap, which is very similar to \cite{marzo2015holographic}.
Single-sided acoustic levitation relies on a phased array to create a stable trap through (i) constructive interference, (ii) acoustic radiation force balancing gravity, and (iii) resistance to external disturbances.

i) Phase calculation for constructive interference: Each transducer emits a wave with a specific phase delay to ensure constructive interference at the levitation point.
This distance-dependent phase delay ($\phi_i$) is calculated by \cite{marzo2015holographic}:
\begin{equation}
    \phi_i = \operatorname{mod}(- k d_i + \sigma, 2\pi)
\end{equation}
where, \begin{itemize}
    \item $k = \frac{2\pi}{\lambda}$ is the wavenumber, where \( \lambda \) is the wavelength of the acoustic wave,
    \item $d_i = \| \mathbf{r} - \mathbf{r}_i \|$ is the distance between the transducer and the levitation point,
    \item $\sigma$ is an additional phase shift to control wavefront shape.
\end{itemize}

In this work, the signature phase shift $\sigma$ is $\pi$ for transducers on one side and $0$ for others, effectively altering the wavefront shape. 
The amplitude factor is ignored in this work, assuming equal energy output, which means that each transducer emits a wave of uniform amplitude.

ii) Acoustic radiation force: The total acoustic field at the levitation point ($P$) is given by:
\begin{equation}
    P(\mathbf{r}) = \sum_{i=1}^{N} e^{j(-k d_i + \sigma)}
\end{equation}

where,
\begin{itemize}
    \item $N$ = 8x8 = 64, represent the total number of transducers in the x- and y-directions, respectively,
    \item $k = \frac{2\pi}{\lambda}$ is the wave number, where \( \lambda \) is the wavelength of the acoustic wave,
    \item $d_i = \| \mathbf{r} - \mathbf{r}_i \|$ is the distance between the transducer and the levitation point,
    \item $\sigma$ is an additional phase shift to control wavefront shape.
\end{itemize}








iii) Stability conditions: For stable levitation, the net acoustic force must balance gravity and any perturbations. 

In this work, we dynamically adjusted the phase control, balanced acoustic forces, and the difference in rotational instabilities, and stable acoustic levitation was achieved, allowing independent contactless transport of objects.

\subsubsection{Independent acoustic robot control system}

Here, we investigate the design of the control system used for the orientation and navigation of the acoustic robot. 
Our acoustic robot utilizes a differential drive system, a simplistic but highly effective mobile robot design that uses two separately driven wheels placed on either side of the robot body.
The kinematic model of the differential drive robot is built upon the relationship between the linear and angular velocities of the robot and the wheel’s angular velocities \cite{klancar2017wheeled}. 
The following equations depict this relationship:
\begin{equation}
\begin{bmatrix}
v \\
{\dot{\theta}} 
\end{bmatrix} = \begin{bmatrix}
\frac{R}{2} & \frac{R}{2} \\
-\frac{R}{L} & \frac{R}{L} 
\end{bmatrix} . \begin{bmatrix}
\omega_R \\
\omega_L 
\end{bmatrix} \label{eq:01}
\end{equation}

where,
\begin{itemize}
    \item $\begin{bmatrix} v & \dot{\theta} \end{bmatrix}$ are the linear and angular velocities of the robot, 
    \item $R$ is the radius of the wheels, and $L$ is the distance between the two wheels,
    \item $\begin{bmatrix} \omega_R & \omega_L \end{bmatrix}$ are the angular velocities vector of the wheels.
\end{itemize}


These equations define the control paradigm of the acoustic robot, enabling precise movements and the effective execution of the desired acoustophoretic functions. 
By regulating wheel speeds, the control system ensures fine-tuned steering, speed, and motion control, which are essential for seamless navigation.
In this work, the acoustic robot control system employs a two-layer proportional control structure, consisting of a position control layer and a final orientation control layer.
Linear and angular velocities are independently evaluated within these layers, but they operate in unison to achieve precise maneuvering. 
While this structure appears to assess linear and angular velocities separately, the robot’s actions remain coordinated. 
Fine-tuning the proportional coefficients was essential to achieve optimal control, ensuring that the robot accurately transitions between points.


\subsubsection{Independent acoustic robot mechanical design}

\begin{figure}[!htb]
    \centering \includegraphics[width=0.95\linewidth]{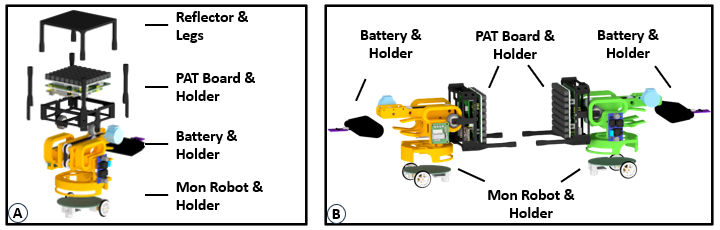}
    \caption{Mechanical design of the contactless object transport system. (A) Exploded view of the independent contactless object transport mechanism, including the reflector component, and (B) Exploded view of the cooperative contactless object transport system, detailing each individual component involved in the operation.}
    \label{MechanicalDesign}
\end{figure}

An acoustic robot is a custom-made robot from modular components (as shown in Figure \ref{MechanicalDesign}A). 
At the base is a Mona robot for mobility\footnote{\url{https://github.com/ICE9-Robotics/MONA\_ESP\_lib/}}, encased in a 3D-printed housing to protect components and allow access to IR sensors.
Above and rear of the Mona robot is a 3D-printed case that holds a Charmast ultracompact battery (5000 mAh, 20 W, USB-C, 77.2 x 35.0 x 24.7 mm, 0.11 kg), powering the phased array. 
Another 3D-printed part secures the phased array on top of the robot.

\subsection{Cooperative Contactless Object Transport}

In this scenario, each acoustic robot is equipped with a phased array that generates precise acoustic fields to manipulate and transport an object in midair. 
Additionally, FPGA and IR-based clock synchronization ensures that both robots coordinate their acoustic forces and maintain stable levitation during contactless object transport.
The primary challenge is to ensure precise force balancing and phase synchronization between the two robots to maintain stable levitation and prevent rotational drift or oscillations while the robots move.
In addition, a leader-follower control system ensures that coordinated backward walks are maintained while maintaining synchronized levitation during contactless cooperative object transport (see Figure \ref{ProposedMethodology}B). 

\subsubsection{Acoustic field generation for synchronized levitation}

The acoustic pressure field $p(x,y,z,t)$ is governed by the wave equation \cite{stevens1899text}:
\begin{equation}
    \nabla^2 p - \frac{1}{c^2} \frac{\partial^2 p}{\partial t^2} = 0
\end{equation}
where, $c$ is the speed of sound in the medium.

For a standing wave field formed by two face-to-face phased arrays, the pressure field can be expressed as:
\begin{equation}
    p(x,t) = P_0 \cos(kx) \cos(\omega t)
\end{equation}
where, $P_0$ is the amplitude, $k$ is the wave number, and $\omega$ is the angular frequency.

Acoustic manipulation relies on the generation of standing waves, which create stable pressure nodes in which objects can be levitated. 
To achieve coordinated transport, we designed a master-slave acoustic field generation algorithm that allows two phased arrays (face-to-face with each other) to interact synchronously (see Figure \ref{ProposedMethodology}).
The master phased array sends a 40 kHz sync signal to synchronize phases, whereas the slave phased array stabilizes and levitates the object by applying the synchronized phases. 


According to Gor'kov \cite{gor1962force}, the total radiation force ($F$) acting on a levitated particle of mass $m$ (sphere shape with a radius $R$) in the standing wave field is:
\begin{equation}
    F = -\nabla U
\end{equation}
where, the Gor'kov potential energy $U$ \cite{gor1962force} is given by:
\begin{equation}
    U = -\frac{1}{2} \rho_0 c^2 \left(\frac{p^2}{\rho_0 c^2}\right)
\end{equation}

In this work, we dynamically adjusted the phase difference between the master- and slave-phased arrays, and stable and synchronized acoustic levitation is achieved, allowing for cooperative contactless transport of objects.

\subsubsection{Leader-follower control system for robot formation}

The leader-follower control system is designed to achieve coordinated backward walking while maintaining synchronized acoustic levitation. 

We define the following terms in the synchronized acoustic levitation and robot formation control system:
\begin{itemize}
    \item Leader acoustic robot ($L$): Generates acoustic forces to push the object forward and provides the sync signal to the follower robot along with the required control output.
    \item Follower acoustic robot ($F$): Receives the sync signal from the leader acoustic robot. Provides synchronized acoustic phases and robot control output to maintain coordinated formation and ensure stable object handling during backward movement.
    \item Object ($O$): A lightweight, rigid body manipulated by the phased array of transducers.
    \item Synchronization offset of the clock ($\Delta t$): Ensures precise clock alignment between the two acoustic phased arrays to maintain stable levitation.
\end{itemize}


The system’s state at time $t$ is defined as:
\begin{equation}
    X(t) = \{ x_L, y_L, \theta_L, x_F, y_F, \theta_F, \Delta t \}
\end{equation}
where,
\begin{itemize}
    \item $(x_L, y_L, \theta_L)$ and $(x_F, y_F, \theta_F)$ are the positions and orientations of the leader and follower acoustic robots, respectively.
    \item $\Delta t$ is the clock synchronization offset between the leader and follower acoustic robots, which must be minimized.
\end{itemize}



\begin{figure}[!htb]
    \centering \includegraphics[width=0.95\linewidth]{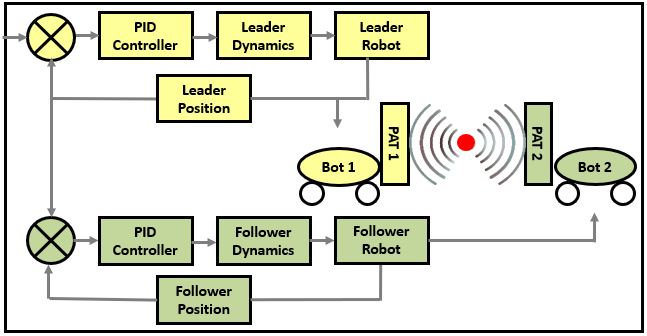}
    \caption{A leader-follower control system for cooperative contactless object transport in mid-air with two acoustic robots.}
    \label{RobotControl}
\end{figure}

The primary goal is to develop a clock-synchronized control strategy and a robot formation control algorithm to ensure stable acoustic levitation and coordinated transport while in motion (see Figure \ref{RobotControl}).

The position of the leader acoustic robot is given by:
\begin{equation}
    x_L(t) = x_0 + v_L t
\end{equation}
where, $x_0$ is the initial position, and $v_L$ is the velocity of the leader acoustic robot.

The follower acoustic robot adjusts its velocity based on the leader acoustic robot's motion using PID controller:

\begin{equation}
\begin{split}
    v_F(t) = v_L(t) + K_p (x_L - x_F) \\
                    + K_i \int (x_L - x_F) dt \\
                    + K_d \frac{d}{dt} (x_L - x_F)
\end{split}
\end{equation}

where, $K_p$ is the proportional gain, which adjusts the follower’s velocity based on the error in the current position, $K_i$ is the integral gain, which accounts for accumulated position error over time to reduce steady-state error, $K_d$ is the derivative gain that predicts future errors and improves stability by reacting to rapid changes.



\begin{figure}[!htb]
    \centering \includegraphics[width=0.95\linewidth]{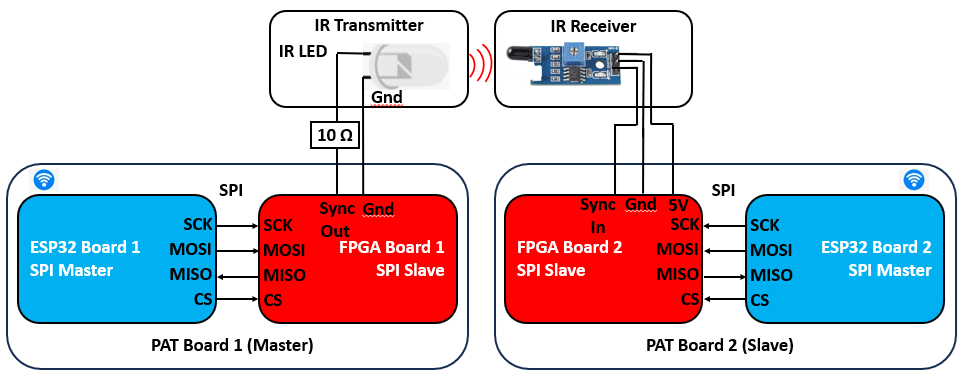}
    \caption{FPGA and IR-based clock synchronization for acoustic field generation.}
    \label{ClockSync}
\end{figure}

This PID-based control ensures that the follower acoustic robot maintains precise synchronization with the leader while minimizing oscillations and delays.
Based on this model, we implemented the following methods:

\begin{itemize}
    \item Leader acoustic robot control: The leader robot determines the trajectory and transmits velocity commands along with the sync signal to the follower robot.
    \item Follower acoustic robot control: The follower robot adjusts its position and orientation based on the commands received while ensuring phase synchronization.
    \item Feedback loop: Both robots utilize odometry and external motion capture data to fine-tune their walking motions and maintain balance.
    \item Clock synchronization: Initially, we tried clock synchronization between the two ESP32 boards (including RTC timer interrupts) using the ESP-NOW protocol over Wi-Fi. Due to Wi-Fi latency, we were unable to synchronize levitation. Hence, we used FPGA and IR-based clock synchronization, in which we achieved the desired result. This clock synchronization ensures that phase-aligned acoustic forces are maintained throughout the cooperative object transport process (see Figure \ref{ClockSync}).
\end{itemize}



By dynamically adjusting the velocity and acoustic phase synchronization of the follower robot based on the leader's movement, coordinated and stable transport is achieved.

\subsubsection{Docking mechanism for cooperative acoustic robots}

We tried to correct the alignment errors between the face-to-face phased array boards with various methods, such as camera-based and motion-capture systems, by taking the position and orientation of the robot and phased array board feedback and correcting it.
However, these methods took time and were unable to resolve the alignment errors.
Hence, we include a mounting mechanism in the modular frame to enable rigid attachments between two acoustic robots. 
We used Neodymium Iron Boron (NdFeB) magnets as passive actuators because of their large strength-to-weight ratio. 
The magnets are round with 3x5mm in height and width. 
The magnets are located on each of the two 3D-printed connectors in the frame (see Figure \ref{MechanicalDesign}B).
These magnets enable two robots to have connections on the two legs, and when two robots are connected face-to-face, the magnets are able to provide a bonding force to align them.
Once attached, undocking is a difficult task and is left to future work.

The above methods are developed in C++ using Microsoft Visual Studio 2022 on Windows 11 with an Nvidia GeForce RTX 3060 Laptop GPU (server and acoustics), C/INO using Arduino IDE 2.3.2, and deployed on Mona robots (ESP32 Wrover board, robot clients) and phased arrays (Adafruit ESP32 Feather V2 boards, acoustic clients). 
VHDL code (acoustics and clock synchronization) is developed using Intel Quartus Prime Lite Edition 18.1 and is deployed on Altera Cyclone IV FPGA boards.
For more information on the methods developed and the results captured, please refer to the ContactlessTransport GitHub page \footnote{\url{https://github.com/narsimlukemsaram/ContactlessTransport/}}. 
  
\section{Experimental Setup, Results and Discussions}

\begin{figure*}[!htb]
    \centering \includegraphics[width=0.95\linewidth]{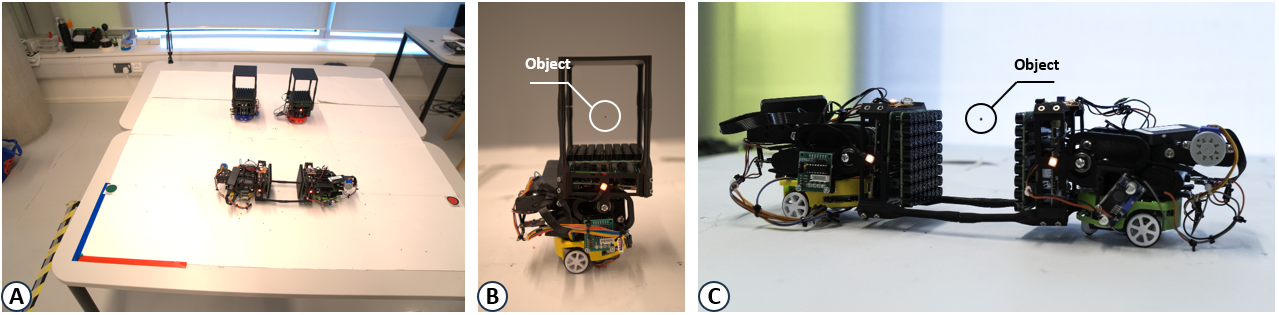}
    \caption{Experimental results of contactless object transport in mid-air using acoustic robots. (A) The 150 × 300 cm test arena used for evaluation, (B) Independent contactless object transport performed by a single acoustic robot, and (C) Cooperative contactless object transport achieved using two acoustically synchronized robots.}
    \label{ExperimentalResults}
\end{figure*}

This section presents the experimental setup and performance evaluation of the proposed contactless object transport system in two scenarios: (i) independent contactless object transport and (ii) cooperative contactless object transport.
The experiments were carried out on a custom-built acoustic robotic research platform within a 150 × 300 cm test arena, using a phase-space tracking system to monitor and analyze the dynamics of object transport (see Figure \ref{ExperimentalResults}A). 


\subsection{\textit{Scenario 1: Independent Contactless Object Transport}}

The experimental setup consists of a single acoustic robot equipped with an 8×8 phased array transducer board (operating at 40 kHz) oriented upward to achieve single-sided acoustic particle levitation for contactless object transport.
The acoustic robot utilizes a high-speed GS-PAT algorithm \cite{plasencia2020gs}, which we modified from the original 16×16 configuration to the 8×8 PAT board, as detailed in Section \ref{Sec_Design}A.
In addition, a modified real-time robotic control algorithm \cite{klancar2017wheeled} governs the motion of the acoustic robot, ensuring precise transport.
A lightweight expanded polystyrene (EPS) particle (1 mm radius) was selected as the testing object due to its suitability for acoustic levitation. 
The experiment also uses a high-speed phase-space\footnote{\url{https://www.phasespace.com/software/}} motion capture and tracking system to monitor the position and orientation of the acoustic robot in real time.



Figure \ref{ExperimentalResults}B shows the results of the independent transport of contactless objects with a single acoustic robot. 
Performance metrics in terms of object levitation stability and transport efficiency are recorded, analyzed, and discussed.
Here are the performance metrics:

i) Levitation stability: Levitation stability is the ability of an object to remain suspended in the air against gravity. 
The phased array of transducers is calibrated to ensure stable acoustic levitation. 
The experiment is evaluated based on the levitation stability strength (pressure) in terms of pascals ($Pa$) using a microphone in the acoustic measurement lab.
In a single-sided levitation scenario, as shown in Table \ref{tab:SimulationVsBeastLab}, the phased array operates at a focal point of 5 cm from the centre of the phased array of transducers board, achieving a simulated pressure of 4469.9 Pa in MATLAB, while microphone measurements recorded 2956.8 Pa in the lab. 
This discrepancy likely stems from practical limitations like acoustic losses, calibration variations, etc.

\begin{figure}[h]
    \centering
    \includegraphics[width=0.90\linewidth]{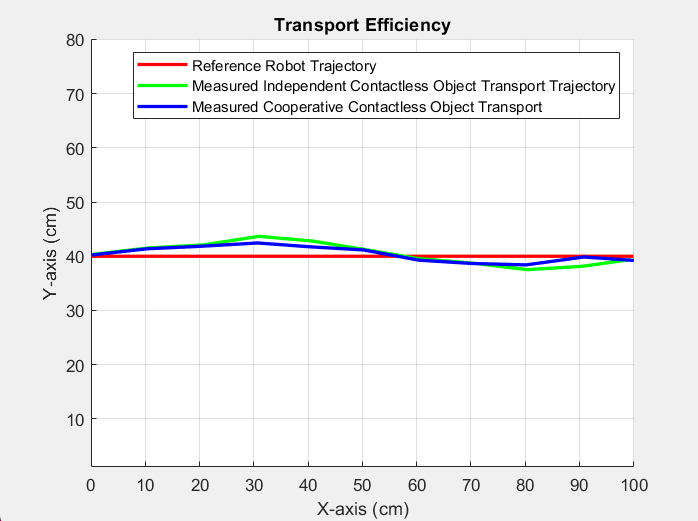}
    \caption{Transport efficiency: Object trajectory during transport.}
    \label{TransportEfficiency}
\end{figure}

\begin{table}[]
\centering
\caption{Levitation Stability: Measurements in MATLAB Simulation vs Measurements in Microphone Lab.}
\label{tab:SimulationVsBeastLab}
\resizebox{\columnwidth}{!}{%
\begin{tabular}{@{}lclc@{}}
\toprule
\multirow{2}{*}{\textbf{\begin{tabular}[c]{@{}l@{}}Simulation in \\ MATLAB/\\ Microphone \\ Measurements\\ in Lab\end{tabular}}} &
  \multicolumn{2}{c}{\textbf{\begin{tabular}[c]{@{}c@{}}Independent \\ (Single-sided levitation)\end{tabular}}} &
  \textbf{\begin{tabular}[c]{@{}c@{}}Cooperative\\ (Face-to-face levitation)\end{tabular}} \\ \cmidrule(l){2-4} 
 &
  \multicolumn{2}{c}{\begin{tabular}[c]{@{}c@{}}Maximum Pressure \\ at Focal Point \\ (in Pascals)\end{tabular}} &
  \begin{tabular}[c]{@{}c@{}}Maximum Pressure \\ at Joint Focal Point\\ (in Pascals)\end{tabular} \\ \midrule
Simulation & \multicolumn{2}{c}{4469.9} & 7106.3 \\ \midrule
Microphone & \multicolumn{2}{c}{2956.8} & 5675.5 \\ \bottomrule
\end{tabular}%
}
\end{table}

ii) Transport efficiency: Transport efficiency is the ability of a transport system to move objects while using less time and resources and having a smaller environmental impact.
The acoustic robot moves forward while carrying the object in the air from the initial position to the final position, and the acoustic forces maintain stability.
We evaluated the performance of the acoustic robot in transporting a 1-mm radius particle at varying velocities ranging from 5 to 10 cm/s.
Due to the weight of the PAT board, the robot was unable to operate effectively at lower velocities (1 to 4 cm/s), and therefore, data for these cases were not captured or analyzed.
However, at a velocity of 5 cm/s, the object was successfully transported with minimal oscillations, as illustrated in Figure \ref{TransportEfficiency}.

\subsection{\textit{Scenario 2: Cooperative Contactless Object Transport}}

The experimental setup consists of two acoustic robots, each equipped with an 8×8 phased array transducer board operating at 40 kHz, arranged in a face-to-face configuration and working in a leader-follower formation to achieve synchronized acoustic levitation for mid-air object transport. 
Both robots utilize a high-speed GS-PAT algorithm \cite{plasencia2020gs}, modified from the original 16×16 configuration to the 8×8 PAT board, as detailed in Section \ref{Sec_Design}B. 
In addition, each robot is governed by a modified real-time robotic control algorithm \cite{klancar2017wheeled} to ensure precise coordination and stability during transport. 
A lightweight expanded polystyrene (EPS) particle (2 mm radius) was selected as the testing object due to its suitability for synchronized acoustic levitation.
The experimental setup incorporates an FPGA and IR-based synchronization mechanism with microsecond precision to maintain phase-aligned acoustic forces, ensuring stable and coordinated levitation.
In addition, a high-speed phase-space motion capture and tracking system is used to continuously monitor the position and orientation of acoustic robots with high accuracy.

\begin{figure}[!htb]
    \centering \includegraphics[width=0.95\linewidth]{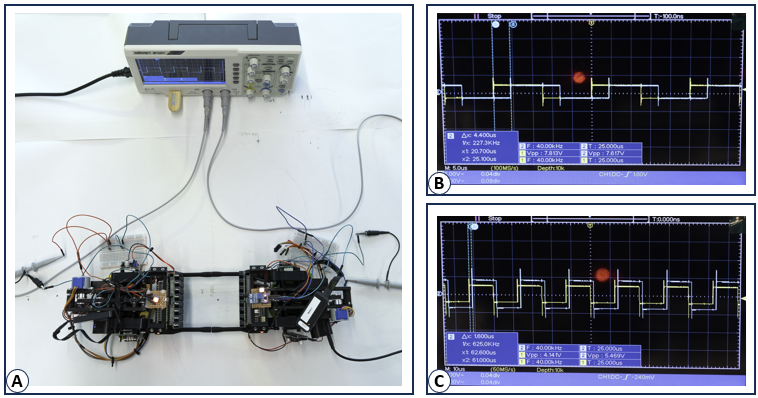}
    \caption{Results of clock synchronization accuracy: (A) Experimental setup of the FPGA and IR-based clock synchronization, (B) Clock synchronization accuracy between two FPGA boards without using IR send-receive communication, and (C) Clock synchronization accuracy between two FPGA boards using IR send-receive communication.}
    \label{ClockSync_Results}
\end{figure}

Figure \ref{ExperimentalResults}C presents the experimental results of cooperative contactless object transport using two acoustic robots. 
The performance of the system is evaluated on the following key metrics:

i) Levitation stability: The phased arrays are calibrated to ensure stable acoustic levitation.
The experiment is evaluated based on the levitation stability strength (pressure) in terms of pascals ($Pa$).
Table \ref{tab:SimulationVsBeastLab} shows the results of the simulation and microphone measurements obtained.
Key insights emerge from these results:
In this face-to-face levitation scenario, joint focus produced higher pressures, with simulated 7106.1 Pa and measured 5675.5 Pa microphone results. This is possibly due to optimized alignment and stronger constructive interference. 
The larger simulation-experiment deviation here implies that higher pressures may amplify nonlinear or environmental damping effects in real measurements.

ii) Transport efficiency:
The leader acoustic robot moves from the initial position to the final position, while the follower acoustic robot dynamically adjusts its acoustic forces to maintain stability and alignment, ensuring synchronized formation during mid-air object transport.
To assess system performance, we conducted experiments transporting a 2-mm radius particle at velocities ranging from 5 to 10 cm/s.
Due to the weight of the PAT board, the robots were unable to operate effectively at lower velocities (1 to 4 cm/s), and data for these cases were not recorded.
However, at velocities of 5 cm/s, the object was successfully transported with minimal oscillations, as shown in Figure \ref{TransportEfficiency}.

iii) Clock synchronization accuracy:
Initially, clock synchronization between two ESP32 boards was tested using the ESP-NOW protocol over Wi-Fi.
However, due to Wi-Fi latency, the achieved synchronization accuracy was 7.8 milliseconds, which was insufficient to maintain stable synchronized levitation.
To overcome this limitation, an FPGA- and IR-based synchronization method was implemented, improving the accuracy to 1.6 microseconds. 
This enhancement enabled reliable and stable acoustic levitation.
Figure \ref{ClockSync_Results} illustrates the synchronization accuracy achieved (A) without using IR-based communication between two FPGA boards and (B) using IR-based communication between two FPGA boards.

\subsection{Discussions}

The proposed independent and cooperative contactless object transport using a single-sided and face-to-face acoustic levitation system demonstrates the capability to transport airborne particles. 
By leveraging constructive interference, the system generates stable acoustic traps that counteract gravity and external disturbances. 
Experimental results validate the effectiveness of the levitation strategy, highlighting its robustness in maintaining particle stability during motion. 
Additionally, the leader-follower synchronization mechanism ensures coordinated transport, minimizing drift and oscillations. 
Although the system successfully achieves contactless transportation, challenges remain in optimizing phase synchronization and force balance for improved transport efficiency. 
Future work will focus on refining acoustic and robotic control algorithms, expanding to multi-particle transportation, and exploring potential applications in material handling, microfluidics, and biomedical fields.

\section{Conclusions}

This work presents a novel acoustic robotic system for the independent and cooperative transport of contactless objects, inspired by biological systems such as ant colonies.
Using phased ultrasonic transducers and robotic control, we generated localized acoustic pressure fields to transport airborne objects with precision.
The experimental results confirm the feasibility of both independent and synchronized cooperative transport, achieving stable levitation with an acoustic pressure of 5675.5 Pascals. Additionally, FPGA- and IR-based clock synchronization attained a precision of 1.6 microseconds.
The system enables smooth object transport at a velocity of 5 cm/sec, validating its effectiveness.
The proposed approach advances acoustophoretic robotics, with potential applications in contactless material handling, micro-assembly, and biomedical fields. 
Future work will explore multi-robot collaboration and dynamic object manipulation for more complex transport tasks.

\addtolength{\textheight}{-12cm}   



\section*{ACKNOWLEDGMENT}

This work was supported by the EPSRC Prosperity partnership program - Swarm Spatial Sound Modulators (EP/V037846/1), and by the Royal Academy of Engineering through their Chairs in Emerging Technology Program (CIET 17/18).
The authors thank Ryuji Hirayama for providing access to the phased array of transducer boards.
The authors also thank Ana Marques for her support in creating the images and video and James Hardwick for his voice-over for the accompanying video.



\bibliographystyle{IEEEtran}
\bibliography{root}

\end{document}